\documentclass[conference]{IEEEtran}
\IEEEoverridecommandlockouts
\usepackage{cite}
\usepackage{amsmath,amssymb,amsfonts}
\usepackage{algorithmic}
\usepackage{graphicx}
\usepackage{textcomp}
\usepackage{xcolor}
\usepackage{dirtytalk}
\usepackage{url}
\usepackage{hyperref}
\usepackage{balance}
\def\BibTeX{{\rm B\kern-.05em{\sc i\kern-.025em b}\kern-.08em
    T\kern-.1667em\lower.7ex\hbox{E}\kern-.125emX}}
\begin{document}
\title{A Multimodal Dataset for Enhancing Industrial Task Monitoring and Engagement Prediction\\
}

\author{\IEEEauthorblockN{Naval Kishore Mehta}
\IEEEauthorblockA{\textit{Academy of Scientific and Innovative Research} \\
\textit{CSIR-Central Electronics Engineering Research Institute}\\
Pilani, India \\
naval.ceeri18a@acsir.res.in}
 
\and
\IEEEauthorblockN{ Arvind }
\IEEEauthorblockA{\textit{Academy of Scientific and Innovative Research} \\
\textit{CSIR-Central Electronics Engineering Research Institute}\\
Pilani, India \\
arvind.ceeri24a@acsir.res.in}
\and
\IEEEauthorblockN{ Himanshu Kumar}
\IEEEauthorblockA{\textit{Academy of Scientific and Innovative Research} \\
\textit{CSIR-Central Electronics Engineering Research Institute}\\
Pilani, India \\
himanshu.ceeri20a@acsir.res.in}
\and
\IEEEauthorblockN{  Abeer Banerjee}
\IEEEauthorblockA{\textit{Academy of Scientific and Innovative Research} \\
\textit{CSIR-Central Electronics Engineering Research Institute}\\
Pilani, India \\
abeer.ceeri20a@acsir.res.in}
\and
\IEEEauthorblockN{  Sumeet Saurav}
\IEEEauthorblockA{\textit{Academy of Scientific and Innovative Research} \\
\textit{CSIR-Central Electronics Engineering Research Institute}\\
Pilani, India \\
sumeet@ceeri.res.in}
\and
\IEEEauthorblockN{ Sanjay Singh}
\IEEEauthorblockA{\textit{Academy of Scientific and Innovative Research} \\
\textit{CSIR-Central Electronics Engineering Research Institute}\\
Pilani, India \\
sanjay@ceeri.res.in}
}

\maketitle

\begin{abstract}

Detecting and interpreting operator actions, engagement, and object interactions in dynamic industrial workflows remains a significant challenge in human-robot collaboration research, especially within complex, real-world environments. Traditional unimodal methods often fall short of capturing the intricacies of these unstructured industrial settings. To address this gap, we present a novel Multimodal Industrial Activity Monitoring (MIAM) dataset that captures realistic assembly and disassembly tasks, facilitating the evaluation of key meta-tasks such as action localization, object interaction, and engagement prediction. The dataset comprises multi-view RGB, depth, and Inertial Measurement Unit (IMU) data collected from 22 sessions, amounting to 290 minutes of untrimmed video, annotated in detail for task performance and operator behavior. Its distinctiveness lies in the integration of multiple data modalities and its emphasis on real-world, untrimmed industrial workflows—key for advancing research in human-robot collaboration and operator monitoring. Additionally, we propose a multimodal network that fuses RGB frames, IMU data, and skeleton sequences to predict engagement levels during industrial tasks. Our approach improves the accuracy of recognizing engagement states, providing a robust solution for monitoring operator performance in dynamic industrial environments. The dataset and code can be accessed from \href{https://github.com/navalkishoremehta95/MIAM/}{https://github.com/navalkishoremehta95/MIAM/}.

\end{abstract}


\begin{figure}[h!]
     \centering
    \includegraphics[scale=.32]{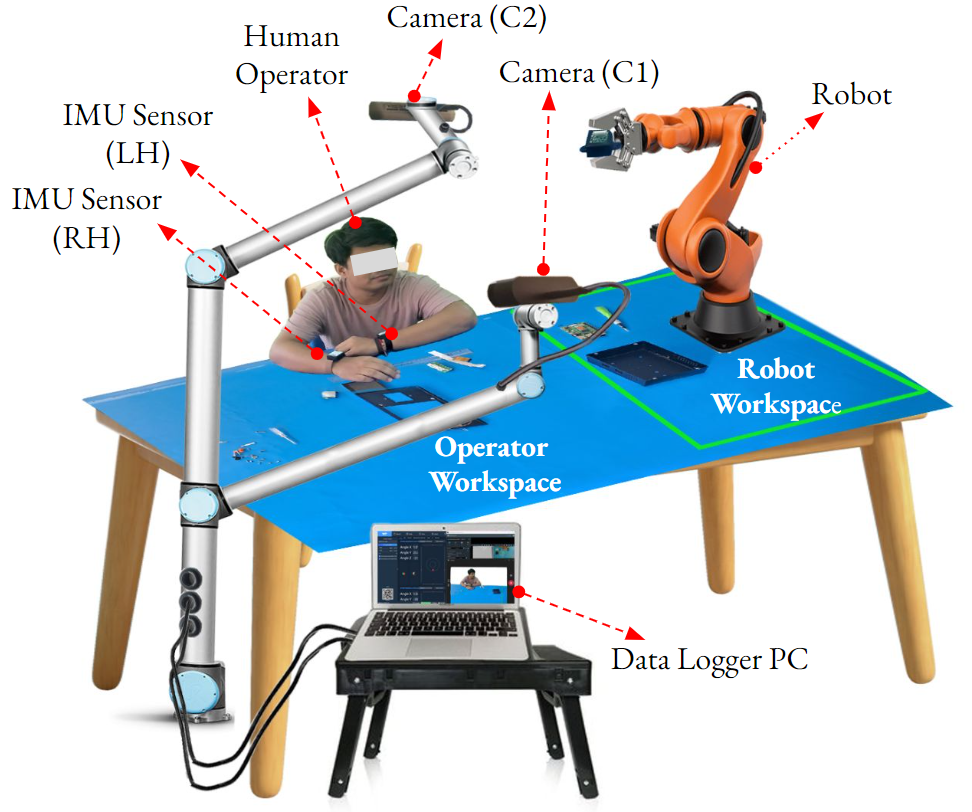}
    \caption{Recording setup for assembly tasks with visual and motion tracking.}
    \label{fig:recording_setup}
\end{figure}

\section{Introduction}
Seamless collaboration between humans and robots in industrial environments is a critical focus in Human-Robot Interaction (HRI) research, especially in dynamic workflows where real-time decision-making, task coordination, and engagement monitoring are essential \cite{d20243d,dobrosovestnova2024we}. Modern industrial settings require sophisticated systems capable of accurately detecting operator actions, monitoring engagement, and managing object interactions to optimize collaboration \cite{R5,mehta2024iar,mehta2024df}. However, existing methodologies, particularly unimodal approaches, often fail to manage the complexities and unpredictability of real-world, unstructured industrial tasks.

Datasets such as Meccano \cite{meccano}, HRI30 \cite{iodice2022hri30}, HA4M \cite{cicirelli2022ha4m} and Enigma \cite{enigma} have contributed significantly by capturing human-object interactions and providing annotated multimodal data for task localization, object recognition, and action classification. While these datasets have advanced research in human-robot collaboration, they still struggle to fully capture the intricate and dynamic nature of industrial workflows where tasks overlap and engagement levels fluctuate. Recent studies underscore the limitations of unimodal approaches, highlighting the need for multimodal frameworks that offer a more comprehensive understanding of human behavior in industrial contexts.

For instance, a study in \cite{R1} demonstrated that integrating audio, visual, and physiological signals outperformed unimodal methods in predicting human engagement in robot-assisted therapy by capturing subtle engagement variations. Another study \cite{R2} showed that incorporating task-related and environmental factors alongside linguistic cues improved disengagement detection in human-robot dialogues. Likewise, \cite{R3} combined physiological signals like heart rate with behavioral observations, enhancing engagement prediction in long-term care. Additionally, \cite{R4} found that integrating visual, auditory, and tactile feedback generated more adaptive and responsive robot behaviors.

The need for robust, real-world datasets capturing the complexity of human-robot interactions in unstructured industrial settings has been emphasized by studies like \cite{R5}. Frameworks such as the one proposed in \cite{R6} illustrate how augmented reality and digital twin systems can improve human-machine collaboration, further driving the need for advanced multimodal datasets. Similarly, the SenseCobot dataset introduced in \cite{R7} highlights the value of integrating sensory data to better understand human-robot dynamics in evolving industrial workflows. Collectively, these studies underline the superiority of multimodal approaches in capturing human engagement dynamics in HRI environments.

To address these challenges, we propose a multimodal pipeline for engagement monitoring using a first-of-its-kind multimodal dataset. Our contributions are summarized below:

\begin{itemize}
    \item We introduce a novel Multimodal Industrial Activity Monitoring (MIAM) dataset designed for industrial settings, focusing on realistic assembly and disassembly tasks. This dataset facilitates the evaluation of key meta-tasks such as action localization, object interaction, and engagement prediction, serving as a valuable resource for advancing research in human-robot collaboration. 
    \item We propose a multimodal network to improve engagement prediction in dynamic environments. By fusing multiple modalities, our approach provides a robust solution for monitoring operator performance, enhancing both engagement detection and overall efficiency in real-world industrial contexts.
\end{itemize}

\begin{table}[h!]
\centering
\footnotesize
\caption{Assembly actions with ideal steps. Abbreviations: CM = Camera Module, DM = Display Module.}
\begin{tabular}{|c|l|c|}
\hline
\textbf{S. No.} & \textbf{Actions (Assembly)} & \textbf{Steps} \\ \hline
1 & take\_rpi\_CM & 1 \\ \hline
2 & take\_rpi\_camera & 1 \\ \hline
3 & align\_rpi\_camera\_with\_CM & 1 \\ \hline
4 & take\_screwdriver & 10 \\ \hline
5 & take\_screw & 13 \\ \hline
6 & screw\_screw\_with\_screwdriver & 8 \\ \hline
7 & put\_rpi\_camera\_CM & 1 \\ \hline
8 & put\_screwdriver & 10 \\ \hline
9 & take\_display & 1 \\ \hline
10 & put\_display & 1 \\ \hline
11 & take\_rpi\_board & 1 \\ \hline
12 & align\_rpi\_board\_on\_display & 1 \\ \hline
13 & take\_fcc\_cable & 2 \\ \hline
14 & plug\_fcc\_cable & 2 \\ \hline
15 & put\_rpi\_board\_DM & 1 \\ \hline
16 & take\_front\_panel & 1 \\ \hline
17 & take\_rpi\_camera\_CM & 1 \\ \hline
18 & align\_rpi\_camera\_CM\_with\_front\_panel & 1 \\ \hline
19 & change\_screwdriver\_bit & 2 \\ \hline
20 & take\_bolt & 4 \\ \hline
21 & take\_nut & 4 \\ \hline
22 & tighten\_nut\_with\_hand & 4 \\ \hline
23 & tighten\_bolt\_with\_screwdriver & 4 \\ \hline
24 & take\_pir\_sensor & 1 \\ \hline
25 & align\_pir\_sensor\_with\_front\_panel & 1 \\ \hline
26 & take\_rpi\_board\_DM & 1 \\ \hline
27 & align\_rpi\_board\_DM\_on\_front\_panel & 1 \\ \hline
28 & take\_display\_mount\_bracket & 2 \\ \hline
29 & plug\_display\_mount\_bracket & 2 \\ \hline
30 & take\_rpi\_hat & 1 \\ \hline
31 & plug\_rpi\_hat & 1 \\ \hline
32 & put\_front\_panel & 1 \\ \hline
33 & take\_back\_panel & 1 \\ \hline
34 & take\_power\_adapter & 1 \\ \hline
35 & tie\_knot\_in\_cable & 1 \\ \hline
36 & plug\_power\_cable & 1 \\ \hline
37 & align\_front\_and\_back\_panel & 1 \\ \hline
38 & put\_complete\_fras & 1 \\ \hline
\textbf{Total} & & \textbf{91} \\ \hline
\end{tabular}
\label{tab:assembly_tasks}
\end{table}

\begin{table}[h!]
\centering
\footnotesize
\caption{Disassembly actions with ideal steps. Abbreviations: CM = Camera Module, DM = Display Module.}
\begin{tabular}{|c|l|c|}
\hline
\textbf{S. No.} & \textbf{Actions (Disassembly)} & \textbf{Steps} \\ \hline
1 & take\_complete\_fras & 1 \\ \hline
2 & unplug\_front\_and\_back\_panel & 1 \\ \hline
3 & unplug\_power\_cable & 1 \\ \hline
4 & put\_power\_adapter & 1 \\ \hline
5 & put\_back\_panel & 1 \\ \hline
6 & take\_screwdriver & 6 \\ \hline
7 & unscrew\_bolt\_with\_screwdriver & 1 \\ \hline
8 & unplug\_rpi\_hat & 1 \\ \hline
9 & put\_rpi\_hat & 1 \\ \hline
10 & unplug\_bolt & 1 \\ \hline
11 & put\_bolt & 2 \\ \hline
12 & put\_nut & 2 \\ \hline
13 & put\_screwdriver & 6 \\ \hline
14 & unplug\_pir\_sensor & 1 \\ \hline
15 & put\_pir\_sensor & 1 \\ \hline
16 & put\_front\_panel & 1 \\ \hline
17 & take\_rpi\_camera\_CM & 1 \\ \hline
18 & unscrew\_screw\_with\_screwdriver & 10 \\ \hline
19 & put\_screw & 10 \\ \hline
20 & unplug\_rpi\_camera\_from\_CM & 1 \\ \hline
21 & put\_rpi\_camera & 1 \\ \hline
22 & put\_CM & 1 \\ \hline
23 & take\_rpi\_board\_DM & 1 \\ \hline
24 & unplug\_fcc\_cable & 1 \\ \hline
25 & put\_fcc\_cable & 1 \\ \hline
26 & unplug\_display\_mount\_bracket & 2 \\ \hline
27 & put\_display\_mount\_bracket & 2 \\ \hline
28 & unplug\_rpi\_board & 1 \\ \hline
29 & put\_rpi\_board & 1 \\ \hline
30 & put\_display & 1 \\ \hline
\textbf{Total} & & \textbf{62} \\ \hline
\end{tabular}
\label{tab:disassembly_tasks}
\end{table}

\section{Methodology}

\begin{figure}[h!]
     \centering
    \includegraphics[scale=.21]{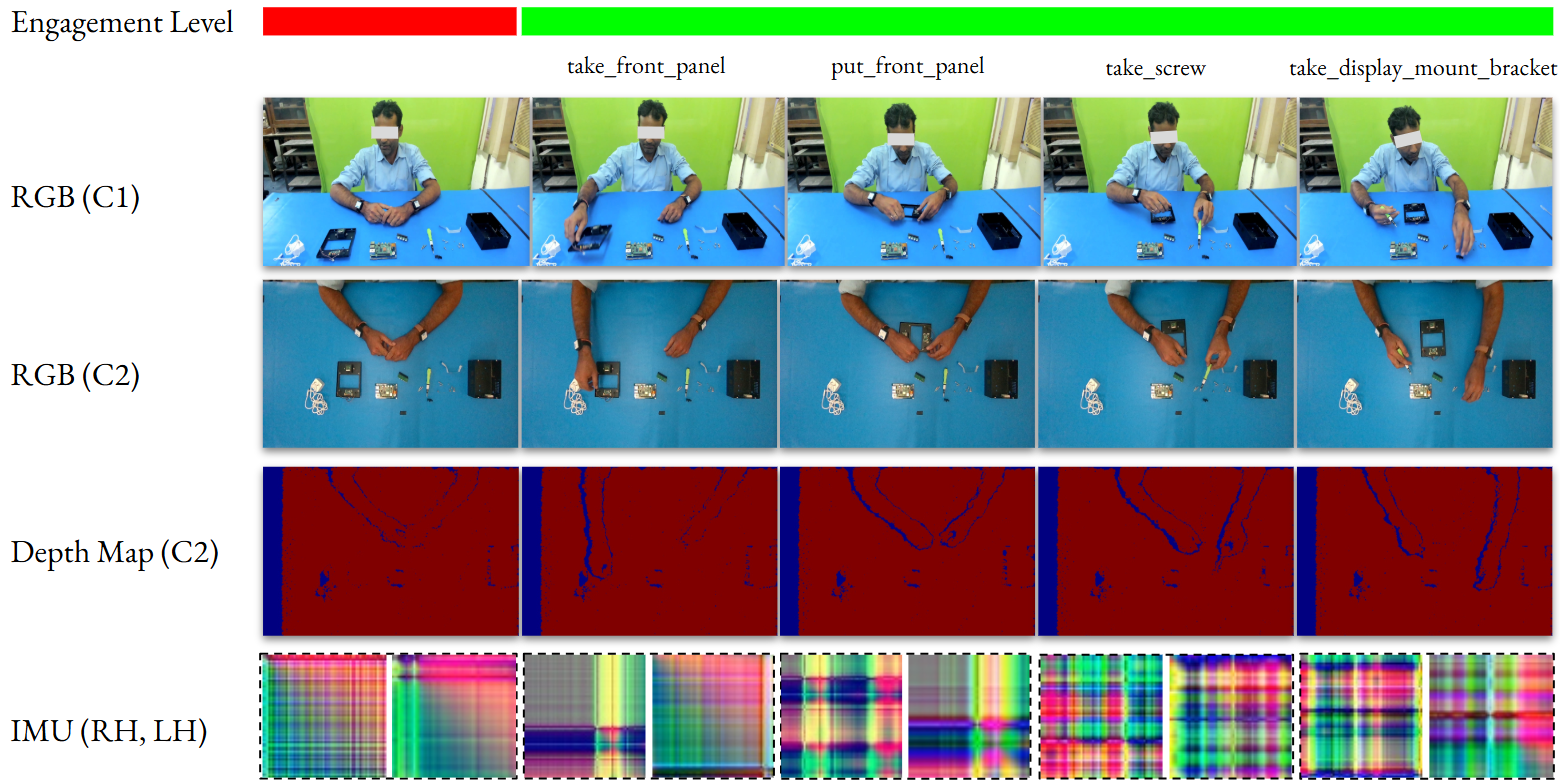}
    \caption{MIAM dataset sample with engagement levels, action labels, RGB and depth views, and IMU data during assembly tasks.}
    \label{fig:samples}
\end{figure}

\subsection{Experiment setup}\label{AA}

The assembly workstation, referred to in Figure~\ref{fig:recording_setup}, is equipped to capture an operator performing tasks using two cameras: a Brio 4K Webcam (C1) for RGB data at \(1920 \times 1080\) resolution at 30 fps, and an Intel RealSense Depth Camera D455 (C2) for RGB-D information. C2 captures RGB data at \(1280 \times 480\) resolution at 30 fps in the RGB8 color format and depth data at \(848 \times 480\) resolution at 30 fps in the Z16 format. WT901BLECL 9-axis BLE Inertial Measurement Unit (IMU) sensors are attached to the operator's left (LH) and right hands (RH) to monitor hand movements. All sensor data, including video and IMU signals, are transmitted to a Data Logger PC for real-time logging and analysis. Camera C1, acting as the master clock, synchronized second-level timestamps with IMU and Camera C2. All devices were linked to a central recording PC for precise, synchronized data capture. The workspace is divided into an Operator Workspace for task execution and a Robot Workspace for collaborative robotic tasks.

\subsection{Data collection}
A total of 22 sessions, spanning over 290 minutes, were recorded by 8 volunteers (75\% male, 25\% female), aged 23 to 37 years (mean age 27, SD 4 years). The tasks were performed in an uncontrolled environment, with participants encouraged to ask questions during the process, though these interactions were less frequent than task engagement. Both engagement and moments of disengagement, such as distractions, asking questions, and taking short breaks, were intentionally captured to reflect diverse scenarios and natural disengagement. The data was collected from industrial assembly and disassembly tasks, where an operator performs a series of steps to construct and deconstruct a Face Recognition-based Attendance System (FRAS) from individual components. These components include the \textit{camera}, \textit{camera module}, \textit{Raspberry Pi (RPI)}, \textit{display}, \textit{PIR sensor}, \textit{screwdriver}, \textit{screws}, \textit{nuts}, \textit{bolts}, \textit{front panel}, \textit{back panel}, \textit{power adapter}, \textit{scale}, \textit{hand tools} (\textit{scissors}, \textit{tweezers}, \textit{wire cutter}), \textit{FCC cable}, \textit{RPI head}, and \textit{display mount bracket}. This process reflects real-world industrial workflows requiring precision, coordination, and the use of multiple tools.  RGB-D camera data was stored in $.bin$ format, RGB video in $.mp4$ format, and both IMU sensor data in $.csv$ format. This process reflects real-world industrial workflows requiring precision, coordination, and the use of multiple tools. Video annotations were performed using the VGG Image Annotator (VIA)\cite{dutta2019via} tool to ensure precise labeling of each step. There were six annotators in total, with three annotating each video independently. Ambiguous cases were marked as "No Label" and later resolved in consensus meetings. Agreement was calculated by averaging scores to ensure consistency.

\begin{figure*}[h!]
     \centering
    \includegraphics[scale=.22]{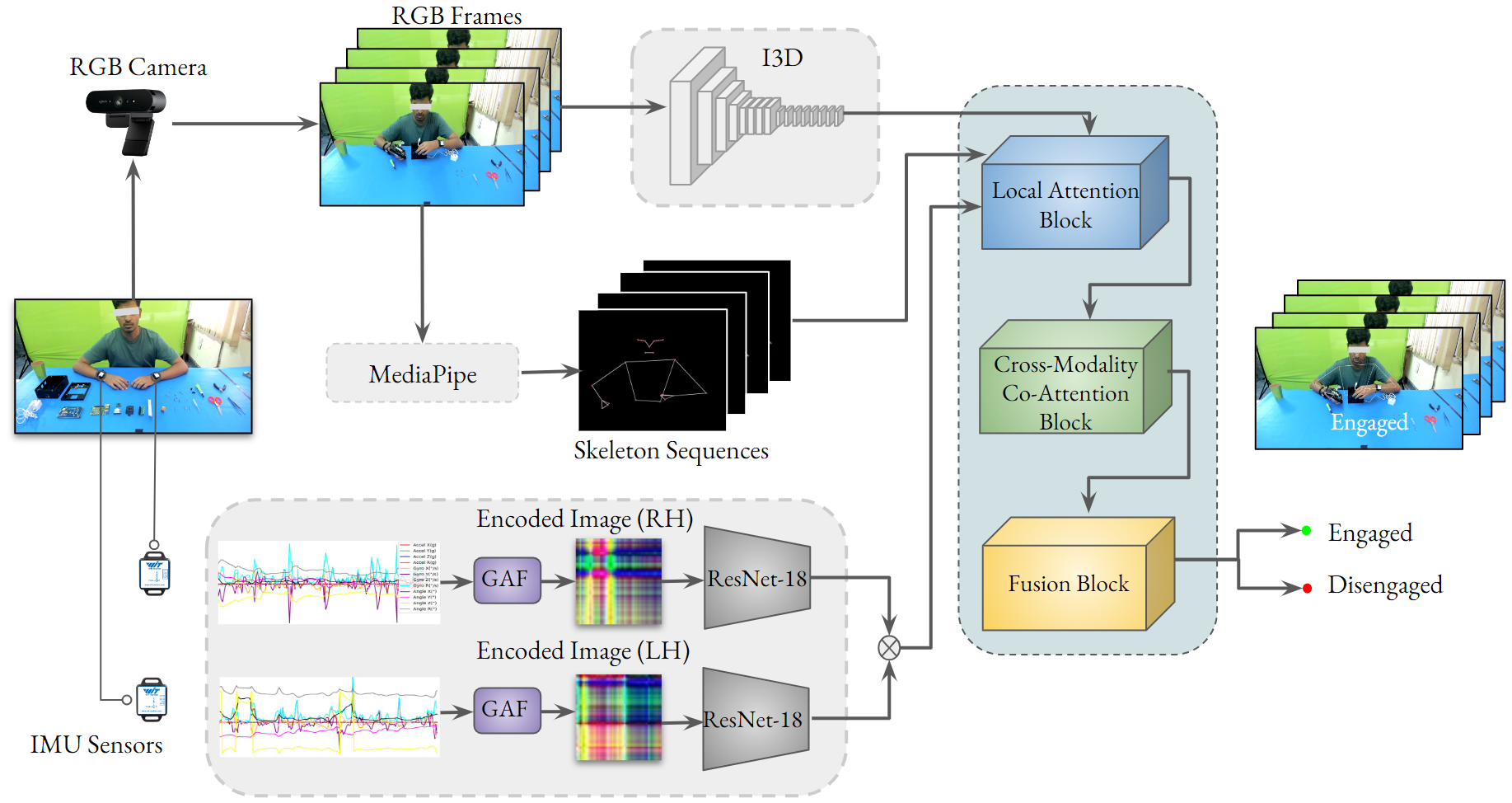}
    \caption{Overview of the multimodal fusion network for engagement recognition, demonstrating the flow from RGB frames, IMU sensors, and pose keypoints through the attention blocks to the final fusion and classification layers.}
    \label{fig:multimodal_network}
\end{figure*}

\subsection{Task and Data Annotation}
This study focuses on three key meta-tasks in the context of industrial assembly and disassembly: action localization, active object localization, and engagement prediction. The MIAM dataset includes time-stamped annotations across multiple modalities, created by multiple annotators. A label quality validation technique was employed to ensure consistency and accuracy across annotators.

Table~\ref{tab:assembly_tasks} \& ~\ref{tab:disassembly_tasks} outlines the 91-step assembly and 62-step disassembly processes. The assembly tasks, involving precise actions like aligning and tightening, require more attention to detail and tool usage, while disassembly focuses on reversal steps like unscrewing and unplugging. Subject bias is evident, with participants showing preferences for handling complex steps differently, particularly in alignment and fastening tasks, which affects performance and task duration. Figure~\ref{fig:samples} illustrates the dataset with multiple modalities. The top row shows engagement levels (green: engaged, red: disengaged) and action labels. The next rows display RGB views: a third-person view (C1) and a top-down view (C2) with RGB and depth data. The final row presents IMU sensor data for the left and right hands (LH, RH), transformed into Gramian Angular Fields (GAF), where the magnitudes of acceleration, angles, and angular velocities \(\left( \sqrt{x^2 + y^2 + z^2} \right)\) were calculated and encoded across RGB channels. 
The dataset is split into training and test sets with a 70:30 ratio at the video level. Each video, uniquely identified by its name, includes corresponding data and labels. Detailed information on file structure, annotations, and access instructions is available on the \href{https://visionai.ceeri.res.in/dataweb/datasets.html}{dataset} page.

\section{Engagement Prediction}
Engagement prediction is critical for improving efficiency and safety in industrial settings. We conducted a preliminary evaluation using the dataset. The training set consists of 1,572 samples (1,441 \textit{Engaged}, 131 \textit{Disengaged}), while the test set contains 386 samples (320 \textit{Engaged}, 66 \textit{Disengaged}).
The proposed multimodal network (as illustrated in Figure~\ref{fig:multimodal_network}) integrates RGB frames, IMU data, and skeleton sequences to predict engagement levels during industrial tasks. The RGB frames are processed via an I3D\cite{carreira2017quo} backbone pre-trained on Kinetic\cite{kay2017kinetics} dataset, while IMU signals are converted into  GAF images for both left and right hands, which are subsequently passed through ResNet-18\cite{he2016deep} encoders pre-trained on ImageNet\cite{deng2009imagenet} dataset. Skeleton sequences, extracted using MediaPipe\cite{lugaresi2019mediapipe}, are embedded via a linear layer. Each modality undergoes intra-modality attention, enhancing modality-specific features, followed by cross-modality co-attention, enabling the model to focus on the complementary information across modalities. The final features from all three modalities are fused using a gated fusion block to predict the operator’s engagement state as either \textit{Engaged} or \textit{Disengaged}.

\begin{table}[h]
\centering
\caption{Engagement recognition accuracy with various modality combinations.}

\begin{tabular}{ll}
\hline
\textbf{Modality}    & \textbf{Accuracy (\%)} \\ \hline
RGB + LH             & 83.20                  \\
RGB + RH             & 83.46                  \\
RGB + LH + RH        & 83.72                  \\
RGB + Pose            & 85.75                  \\  
RGB + LH + RH + Pose & 86.79                  \\ \hline

\label{accuracies}
 
\end{tabular}
\end{table}

\section{Results and  Discussion}

To address the research questions outlined in the preceding section, we evaluated the effect of different data modalities on engagement prediction accuracy.
Table~\ref{accuracies} shows the progressive improvement in accuracy with added modalities. RGB + LH achieves 83.20\%, while RGB + RH provides a slight 0.31\% increase to 83.46\%. Combining both hands (RGB + LH + RH) further improves accuracy by 0.26\%, reaching 83.72\%, indicating the benefit of bilateral movements. Introducing pose data with RGB alone (RGB + Pose) achieves a significant improvement, raising accuracy to 85.75\%. The largest gain, 3.67\%, is achieved with the addition of pose data to hand-based modalities (RGB + LH + RH + Pose), raising accuracy to 86.79\%. The full-body context provided by pose data significantly enhances engagement prediction, with cross-modality attention effectively integrating complementary features.

The dataset is unique in its multimodal nature, integrating multi-view RGB, depth, and IMU data from real-world industrial workflows. It addresses key challenges in industrial assembly, such as accurately capturing operator behavior in dynamic, noisy environments. By fusing data from multiple sources, this dataset enables more precise predictions of engagement, action localization, and object interaction. Its real-world relevance makes it an essential tool for advancing HRI research. This dataset not only provides a platform for robust engagement prediction but also supports broader research in action localization and object interaction in complex industrial settings. The dataset is stored on our institution’s server and will be made available under the CC BY 4.0 license.

\section{conclusion}
This paper presents a multimodal dataset designed to overcome the challenges of monitoring operator actions in industrial settings. By capturing real-world assembly and disassembly tasks with multi-view RGB, depth, and IMU data, the dataset provides a foundation for evaluating key meta-tasks such as action localization, object interaction, and engagement prediction. The combination of diverse data modalities offers a comprehensive resource for studying complex workflows and operator behavior in dynamic environments. We also proposed an engagement prediction network, which significantly improved accuracy, particularly with the integration of pose data, underscoring the importance of full-body context. Future work will focus on integrating real-time feedback systems to enhance operator performance and safety.

\section*{Acknowledgment}
The authors thank the Director of CSIR-CEERI for supporting AI research, the volunteers for their contributions to the database, and acknowledge funding from CSIR-HRDG (Senior Research Fellowship) and MeitY (Visvesvaraya Fellowship), Government of India.

\bibliographystyle{IEEEtran}
\balance
\bibliography{ref}

\end{document}